# Contour Detection from Deep Patch-level Boundary Prediction

Teck Wee Chua    Li Shen
Institute for Infocomm Research, A-Star
Singapore
tewchua, lshen@i2r.a-star.edu.sg

*Abstract*— In this paper, we present a novel approach for contour detection with Convolutional Neural Networks. A multi-scale CNN learning framework is designed to automatically learn the most relevant features for contour patch detection. Our method uses patch-level measurements to create contour maps with overlapping patches. We show the proposed CNN is able to to detect large-scale contours in an image efficiently. We further propose a guided filtering method to refine the contour maps produced from large-scale contours. Experimental results on the major contour benchmark databases demonstrate the effectiveness of the proposed technique. We show our method can achieve good detection of both fine-scale and large-scale contours.

## I. INTRODUCTION

Contour detection has long been a core problem in computer vision. Such contours mark the boundary between physically separated objects and provide important cues of low and high-level understanding of scene content. Contour detection is a critical preprocessing step for a variety of tasks, including segmentation, object detection and scene understanding. High-quality image segmentation and object proposals have increasingly been relying on contour analysis. Contours and segmentations have also seen extensive uses in shape matching and object recognition.

Despite its importance and long tradition, accurately finding contours in natural images is still a challenging problem. The main difficulty is that visually salient edges correspond to a large variety of visual phenomena. Motivated by this observation, recent papers [17, 1, 6, 14, 15, 20, 18, 19, 11] have explored the use of learning for boundary detection. These approaches take an image patch and predict the presence/absence of a boundary at the center pixel of the patch. One of the challenges in accurate boundary detection is the seemingly inherent contradiction between the correctness of differentiating a boundary edge and precision of localizing the boundary [10]. In contrast to image edges detection, contour detection aims to detect objects' boundary which require more global context. As a result, such detectors tend to use relatively large input patches, thus leading to blurry contour. Moreover, such per-pixel classifiers are limited by their locality since they treat each pixel independently. This often leads to noisy and broken contours. Hence, global optimization framework such as MRF and CRF is used as post-processing step to enforce local contour consistency and smoothness over neighboring labels. As shown in [14, 7], edges in a local patch are actually highly interdependent. Recently, [7, 16, 9] proposed to predict patches' structured labels. Such methods require the learning of edge dictionary. The prediction performance is highly depending on how good the edge dictionary can approximate the input image. In this paper, we propose a novel CNN based framework for contour detection. Given an input image patch, we compute the likelihood that the image patch contains a boundary. A multi-scale CNN learning framework is designed to automatically learn the most relevant features for contour patch detection. In contrast to contour detection at center pixels, our patch-based boundary prediction is spatially invariant. i.e. as long as the input patch contains sufficiently strong boundary, we consider the patch as a boundary patch regardless the location of the boundary in the patch. Instead of independently assigning a class label to each pixel, our patch-level measurements create contour maps with overlapping patches that enforcing interdependent decisions. Our CNN achieves robustness by combining the neighboring prediction results. We also show that by using multi-scale inputs that capture local and global information, the CNN is able to harness all relevant information to learn objects' contour effectively with only simple network architecture.

We further propose a guiding filtering method to refine the contour maps. Specifically, we construct the guided filters for each local patch using the predicted large-scale contour. The guiding filter is then applied to the corresponding local patch in gradient domain to "select" the small-scales contours which align to the large-scale contours of the patch. Experimental results show that our method can achieve good detection of both fine-scale and large-scale contours.

## II. RELATED WORK

Early work [3] to edge detection focus on the detection of intensity or color gradients. More recent local approaches [17, 13, 12, 1, 18, 10] take into account color and texture information and explores edge detection under more challenging conditions and make use of learning techniques for cue combination. Recently, some data-driven learning approaches have been developed for boundary detection. Dollar et al. [6] learn an edge classifier in the form of probabilistic boosting tree from thousands of simple features computed on image patches. Mairal et al. [15], and Ren et al.

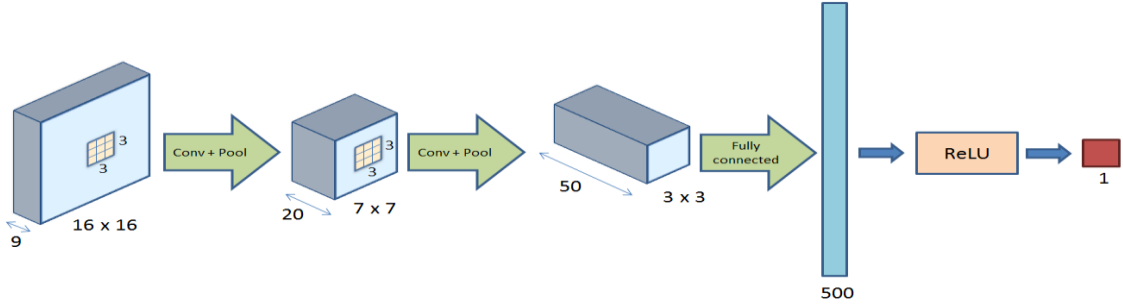

Figure 1. CNN architecture used for contour learning. The network takes in $16\times16\times9$ multi-channel image patch as input and the input. Note that although the output is $1\times1$ we augment it to $16\times16$ before performing simple voting with neighboring patches.

[19] use representations based on sparse coding to learn a discriminative dictionary of contour patches. [19] demonstrates that contour detection can be vastly improved by replacing the hand-designed features of [1] with rich representations that are automatically learned from data. Lim et al. [14] propose an edge detection approach that classifies edge patches into sketch tokens using random forest classifiers. All the approaches are pixel-level prediction, which compute edges independently at each pixel given its surrounding image patch. The per-pixel outputs are inherently noisy with poor contour continuity. To overcome this, the predicted posteriors are usually fed to a global optimization step to enforce label consistency and continuity of contours. Global methods utilize local measurements and embed them into a framework which minimizes global cost over all disjoint pairs of patches. For example, [1, 2, 10] rely on some form of spectral methods while [20, 14] use Conditional Random Field (CRF) framework to enforce curvilinear continuity of contours. Zhu et al. [22] use circular embedding to enforce orderings of edges. Edges in a local patch are highly interdependent [14]. Dollar and Zitnick [7] predict local edge masks in a structured learning framework applied to random decision forests. They predict the labels of multiple pixels in a local patch simultaneously. [7] can recover better local edge structures (local consistency), and avoid assigning implausible label transitions. The structured decision tree framework learns the leaf labels as part of the tree training. Recently, there have also been attempts to apply deep learning methods to the task of contour detection. Marire et al. [16] train the using sparse coding approach. They encode an input image using the generic dictionaries and then reconstruct using the transfer function. N4 fields [9] rely on dictionary learning and the use of the Nearest Neighbor algorithm within a CNN framework. Both [16] and [9] predict the boundaries of a whole patch. Kivinen et al. uses a two-stream CNN architecture to compute edges at each pixel given its surrounding patch. We train our CNN to detect boundary patches. In contrast to contour detection at center pixels, our patch-based boundary prediction is spatially invariant. We do not need precisely predict boundary at pixels. And, using patch-level measurements with overlapping patches allow us to make interdependent predictions. That is why our CNN is efficient and robust.

## III. COARSE BOUNDRY DETECTION USING CNN

The proposed network is shown in Fig. 1. The network consists of two alternating convolutional and max-pooling layers, followed by a Rectified Linear Units (ReLU) layer and finally two fully connected layers. By feeding multiscale inputs to a single CNN that shares the same set of convolutional filters, the network will be able to learn the underlying relationship between local and global contour cues. Finally, we employ simple voting scheme to combine multiple predictions from overlapping $16\times16$ patches.

### A. Implementation Detials

In order to train our network, we first generate the average of ground truth maps by multiple human labelers and normalized to [0, 1]. Stronger edges indicate that more labelers agree on the presence of contour and vice-versa. We apply grid sampling of $16\times16$ patches on the averaged edge map and sum the total edge intensity in the patches. Note that in contrast to many other works, we do not rely solely on central pixel contour intensity for sample selection. This is because the ground truth maps generated by different human labelers are not aligned. By considering all pixels in a local patch, we can ensure that most true contour edges are sampled. Positive patches are selected when the sums exceed threshold of 10 and all other patches are considered as negative patches. Since positive samples are generally much lesser than negative patches, we set the number of positive samples as the upper-bound of the number of negative samples. We randomly select negative samples to match the upper-bound number. Such negative samples may include homogeneous patches and patches with non-contour edges (hard negatives). By having equal proportion of positive and negative samples, we aim to reduce training bias in the convolutional network. In total, we extracted $186842$ and $90248$ patches from training and validation images respectively. For each location, we extract three color patches with sizes of $16\times16$, $32\times32$, and $64\times64$, then resize them to $16\times16$. After that, the patches are stacked to form a 9-channel input.

Our structured CNN achieves robust results by combining the neighboring prediction results. Each pixel collects class hypotheses from the patch-level labels predicted for itself and neighboring patches. We employ a simple spatial averaging scheme to combine the overlapping predictions.

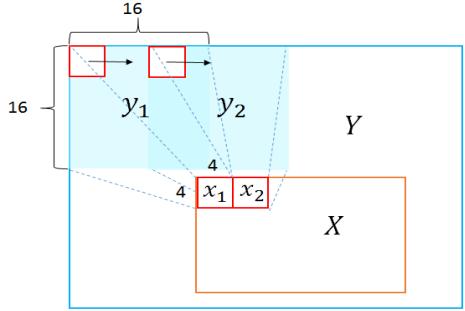

Figure 2. Illustration of refinement. $X$ is the $1/4$ down sampling detection results of our CNN. Y is the gradient of the input image.

## IV. CONTOUR REFINEMENT

We now describe the refinement to our method to generate the fine-scale contours. The coarse contour detection results actually give good prediction of the objects' outline. Given coarse contours, we want to recover the fine-scale contours from the input image.

To address this problem, we introduce a simple guided filtering procedure. Our core observation is that intuitively, the fine-scale contours should have similar structure of the large-scale contours at the same location. Therefore, we can use the predicted large-scale contours as a token to select the most matching local contours.

The refinement takes a down sampled rough contour prediction $x \in X$ and the gradient of the input image $y \in Y$ and produces the fine-scale contours that aligns to $x$. Let $x_i$ denote the contour patch, $y_i$ denote the corresponding gradient patch of the input image. We use $x_i$ as a filter, and apply it on $y_i$ to compute $\tilde{y}_i(j) = \int y_i(l) x_i(j-l)$. For the overlapping region $\Omega_i = y_i \cap y_{i+1}$, we compute $\tilde{y}_i(j)$ as $\max(\tilde{y}_i(j), \tilde{y}_{i+1}(j))$ where $j \in \Omega_i$. Let $X^m$ denote the $2^{-n}$ down sampling contour map, and $Y^n$ denote the filtered results with $X^m$. We can employ the multi-scale scheme as $c(j) = \sum_n \tilde{y}_i^n(j)$. In our experiment, a single scale usually is enough. Fig. 3 show our refinement result with $n = 2$ filtering results.

## V. EXPERIMENTAL RESULTS

We evaluate our contour detector on the BSDS500 dataset. The dataset contains 200 training, 100 validation, and 200 testing images. The performance of the detector is evaluated using three standard measures: fixed contour threshold (ODS) per-image best threshold (OIS) and average precision (AP).

We analyze the performance of single scale and multiple scales inputs. In particular, we test our algorithm using single input scale $16 \times 16$ or $32 \times 32$ or $64 \times 64$ and combination of all three scales. As shown in Fig. 4, CNN contour output from $16 \times 16$ scale is inherently noisy. The detector is able to capture many small edge details other than objects contours. The detector output from $32 \times 32$ scale is less noisy as the CNN is able to detect objects contour better while ignoring some non-contour edges; however, the classification at large textured regions such as brick wall and vegetation is still erroneous. CNN with $64 \times 64$ scale emphasizes more on large objects' contour but the side effect is that small true positives are being removed at the same time due to macro level information processing. In the experiments, we observe that smaller scale input is good at capturing small edges (high recall) including non-contour edges. On the other hand, larger scale input allows CNN to learn objects contour at macro level (high precision) while sacrificing the capability to detect small objects contour. Tab. 1 shows the progressive improvement in all three metrics when using larger single patch size. Combining all three input scales produces significantly better performances.

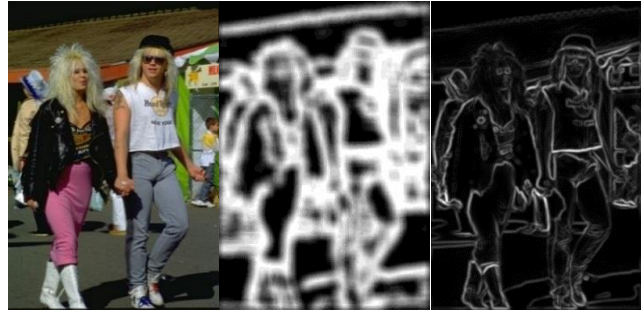

Figure 3. Prediction of fine-scale contours. Left: input image. Middle: $16 \times 16$ CNN output. Right: refinement result.

TABLE I. COMPARISON OF SINGLE SCALE AND MULTI-SCALE PATCHES

| Patch Scale | ODS | OIS | AP |
|---|---|---|---|
| 16 | 0.69 | 0.74 | 0.70 |
| 32 | 0.71 | 0.75 | 0.71 |
| 64 | 0.71 | 0.75 | 0.73 |
| 16, 32, 64 | 0.75 | 0.78 | 0.78 |

TABLE II. PERFORMANCE COMPARISON WITH DIFFERENT CNN LAYERS

| Architecture | ODS | OIS | AP |
|---|---|---|---|
| 2 layers(Conv1-20, conv2-50) | 0.75 | 0.78 | 0.78 |
| 3 Layers (Conv1-30, Conv2-60, Conv3-60) | 0.74 | 0.77 | 0.76 |

TABLE III. PERFORMANCE BENCHMARK ON BSDS500 DATASET

| Methods | ODS | OIS | AP |
|---|---|---|---|
| Canny | 0.60 | 0.63 | 0.58 |
| Felz-Hutt[8] | 0.61 | 0.64 | 0.56 |
| Normalized Cuts[5] | 0.64 | 0.68 | 0.45 |
| Mean Shift[4] | 0.64 | 0.68 | 0.56 |
| Gb[13] | 0.69 | 0.72 | 0.72 |
| ISCRA[21] | 0.72 | 0.75 | 0.46 |
| gPb-owt-ucm[1] | 0.73 | 0.76 | 0.73 |
| Sketch Token[14] | 0.73 | 0.75 | 0.78 |
| DeepNet[11] | 0.74 | 0.76 | 0.76 |
| SCG[19] | 0.74 | 0.76 | 0.76 |
| SE[7] | **0.75** | 0.77 | **0.80** |
| **Our Method** | **0.75** | **0.78** | 0.78 |

Tab. 2 shows the performance comparison with different number of CNN layers. There is a slight performance drop when using more convolutional layers. One explanation is that higher convolutional layer tends to learn the outline of large objects but at the expense of sacrificing finer contour details. While 3-layer CNN provides cleaner contour results,

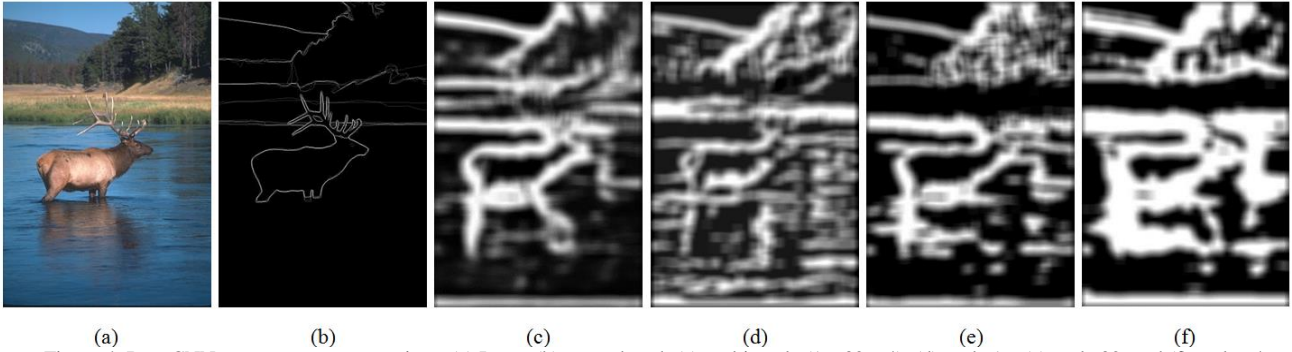
Figure 4. Raw CNN contour maps comparison. (a) Input, (b) ground truth (c) multi-scale (16, 32, 64), (d) scale 16, (e) scale 32, and (f) scale 64

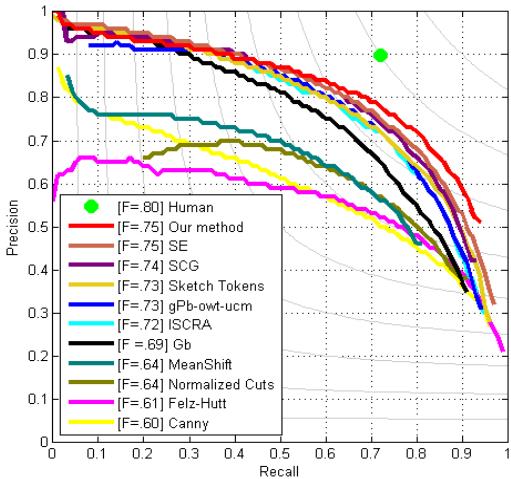
Figure 5. Precision-Recall curve on BSDS500 dataset

there are also more true positives at smaller contour area not being detected.

Fig.6 shows the decomposition results of the compared methods in the BSDS500 dataset. Our method successfully differentiate the object contour edges from shading and texture edges. Tab. 3 shows the performance of our methods compared with the previous methods in the BSDS500 dataset. The Precision-Recall curves are shown in Fig. 5. The proposed method generally has the best performance.

## VI. CONCLUSION

In this paper, we have proposed a simple but efficient CNN-based framework for contour detection. We demonstrated that robust contour detection can be achieved by only using simple CNN network with shallow architecture. The coarse contour predictions are obtained from the patch-based boundary detection results by using a multi-scale CNN. The multi-scale framework allows CNN to learn the latent relationship between local and global contour cues. The fine-scale contours are then recovered by guided filtering. Experimental results show that our method achieves state-of-the-art results on the most popular BSDS500 dataset.

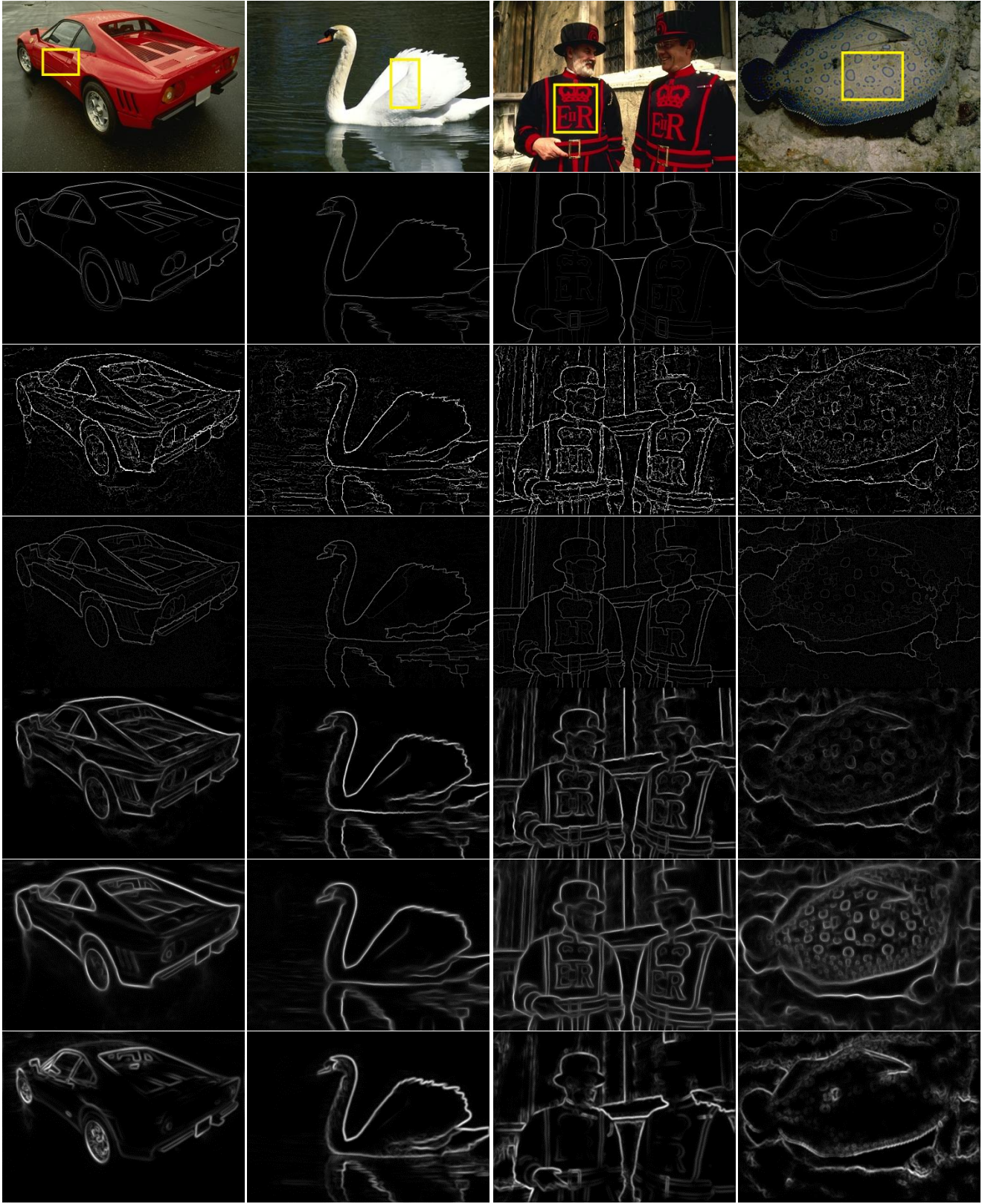

Figure 6. Comparison of contour detection. From top to bottom: input, ground truth, Sketch Tokens [14], MCG [2], SE [7], N4 [9], and our proposed method. Our method can correctly differentiate the object contours from shading edges and texture edges as shown in the yellow regions.